\documentclass{article}
\usepackage{arxiv}
\usepackage[utf8]{inputenc} 
\usepackage[T1]{fontenc}   
\PassOptionsToPackage{hyphens}{url}
\usepackage[pagebackref=true,breaklinks=true,letterpaper=true,colorlinks,bookmarks=false,citecolor=teal]{hyperref}
\usepackage{url}        
\usepackage{booktabs}   
\usepackage{amsfonts}     
\usepackage{nicefrac}  
\usepackage{microtype}   
\usepackage[dvipsnames, svgnames, x11names]{xcolor} 
\usepackage{tcolorbox}
\usepackage{graphicx}
\usepackage{bbding}
\usepackage{subcaption}
\usepackage{wrapfig}
\usepackage{bm}
\usepackage{listings}
\usepackage[ruled,vlined]{algorithm2e}

\usepackage{amsmath}
\usepackage{amssymb}
\usepackage{mathtools}
\usepackage{amsthm}
\usepackage{multirow}
\usepackage[capitalize,noabbrev]{cleveref}
\theoremstyle{plain}

\theoremstyle{definition}

\theoremstyle{remark}

\usepackage{accents}
\usepackage[textsize=tiny]{todonotes}

\usepackage[utf8]{inputenc}
\usepackage[english]{babel}
\usepackage{mathrsfs}
\usepackage{times}
\usepackage{latexsym}
\usepackage{comment}
\usepackage{pifont}
\usepackage{lipsum}

\usepackage[]{mathtools}   
\def\$#1\${\begin{align*}#1\end{align*}}
\usepackage{enumitem}
\usepackage[]{amsthm} 
\usepackage[]{amssymb}
\usepackage{bbm}
\usepackage{xcolor}
\usepackage{colortbl}
\definecolor{best}{HTML}{BAFFCD}
\definecolor{issue}{HTML}{FFC8BA}
\definecolor{bad}{HTML}{FFC87C}

\usepackage{amsmath,amsfonts,bm,amsthm,dsfont}

\def\eqref#1{equation~\ref{#1}}

\def\1{\bm{1}}
\def\0{\bm{0}}

\DeclareMathAlphabet{\mathsfit}{\encodingdefault}{\sfdefault}{m}{sl}
\SetMathAlphabet{\mathsfit}{bold}{\encodingdefault}{\sfdefault}{bx}{n}

\usepackage{tcolorbox}
\definecolor{grey}{rgb}{0.33, 0.33, 0.33}

\newcommand{\squishlist}{
\begin{list}{{{\small{$\bullet$}}}}
{\setlength{\itemsep}{1pt}      \setlength{\parsep}{0pt}
\setlength{\topsep}{-2pt}       \setlength{\partopsep}{0pt}
\setlength{\leftmargin}{1em} \setlength{\labelwidth}{1em}
\setlength{\labelsep}{0.5em} } }
\newcommand{\squishend}{  \end{list}  }

\makeatletter
\renewcommand*\env@matrix[1][*\c@MaxMatrixCols c]{%
  \hskip -\arraycolsep
  \let\@ifnextchar\new@ifnextchar
  \array{#1}}
\makeatother

\def\tr{\mathop{\text{tr}}\kern.2ex}

\long\def\comment#1{}

\def\tr{\mathop{\text{Tr}}}

\newcommand{\bel}{\begin{eqnarray}\label}
\newcommand{\eel}{\end{eqnarray}}
\newcommand{\bes}{\begin{eqnarray*}}
\newcommand{\ees}{\end{eqnarray*}}

\usepackage{xcolor}

\ifodd 1

\newcommand{\clar}[1]{\textbf{\color{green}(NEED CLARIFICATION: #1)}}
\newcommand{\response}[1]{\textbf{\color{magenta}(RESPONSE: #1)}}
\else

\newcommand{\com}[1]{}
\newcommand{\clar}[1]{}
\newcommand{\response}[1]{}
\fi
\newcommand{\RNum}[1]{\uppercase\expandafter{\romannumeral #1\relax}}
\newcommand{\methodname}{TAPTR}
\newcommand{\fulltitle}{\methodname: Tracking Any Point with Transformers as Detection}
\title{\fulltitle} 

\stepcounter{footnote}  

\author{
    \textbf{Hongyang Li}$^{1,2}$\qquad
    \textbf{Hao Zhang}$^{2,3}$  \qquad
    \textbf{Shilong Liu}$^{2,4}$  \qquad 
    \textbf{Zhaoyang Zeng}$^{2}$ \\
    \textbf{Tianhe Ren}$^{2}$\qquad
    \textbf{Feng Li}$^{2,3}$\qquad
    \textbf{Lei Zhang}$^{1,2}$\thanks{Corresponding author.} \\
    $^1$South China University of Technology. \\
    $^2$International Digital Economy Academy (IDEA). \\
    $^3$The Hong Kong University of Science and Technology. \\
    $^4$Dept. of CST., BNRist Center, Institute for AI, Tsinghua University. \\
    \vspace{0.9em}
    \centerline{\url{taptr.github.io}}
}

\begin{document}

\footnotetext{This work was done while Hongyang Li, Hao Zhang, Shilong Liu, and Feng Li were interns at IDEA.}

\maketitle

\newcommand{\myPara}[1]{\vspace{.05in}\noindent\textbf{#1}}

\begin{abstract}
In this paper, we propose a simple and strong framework for Tracking Any Point with TRansformers ({\methodname}). Based on the observation that point tracking bears a great resemblance to object detection and tracking, we borrow designs from DETR-like algorithms to address the task of TAP. In the proposed framework, in each video frame, each tracking point is represented as a point query, which consists of a positional part and a content part. As in DETR, each query (its position and content feature) is naturally updated layer by layer. Its visibility is predicted by its updated content feature. Queries belonging to the same tracking point can exchange information through self-attention along the temporal dimension. As all such operations are well-designed in DETR-like algorithms, the model is conceptually very simple. We also adopt some useful designs such as cost volume from optical flow models and develop simple designs to provide long temporal information while mitigating the feature drifting issue. Our framework demonstrates strong performance with state-of-the-art performance on various TAP datasets with faster inference speed. 

\end{abstract}
\section{Introduction}

Understanding every pixel in a video and tracking their motions is a fundamental task in computer vision, which is of great importance to video object tracking, segmentation, action recognition, and physical world understanding. 
Previously, this task is normally simplified to optical flow estimation and has received much attention~\cite{teed2020raft, Xu_Ranftl_Koltun_2017, Sun_Yang_Liu_Kautz_2018, Wang_Zhong_Dai_Zhang_Ji_Li_2020, Jiang_Lu_Li_Hartley_2021, Huang_Shi_Zhang_Wang_Cheung_Qin_Dai_Li, Shi_Huang_Li_Zhang_Cheung_See_Qin_Dai_Li_2023, Zhao_Zhao_Zhang_Zhou_Metaxas}.
Since optical flow mainly solves the correspondence problem between two consecutive frames, the lack of long-range temporal information makes it ineffective in handling the situation when a tracking point is occluded. The works~\cite{vendrow2023jrdb, Guler_Neverova_Kokkinos_2018, wang2020combining, ning2020lighttrack, sun2022dancetrack} dedicated to semantic key point tracking can handle the problem of occlusion. However, the semantics of tracking targets are limited to only a small range such as the joints of humans. To break the limitation of optical-flow estimation and key-point tracking, some recent works~\cite{neoral2024mft, harley2022particle, zheng2023pointodyssey, doersch2022tap, doersch2023tapir, karaev2023cotracker} propose to track any arbitrary point specified by a user in the whole video and formalize the task as Tracking Any Point (TAP).

\begin{figure*}[t]
    \vspace{-0.1mm}
    \centering
        \includegraphics[width=1\linewidth]{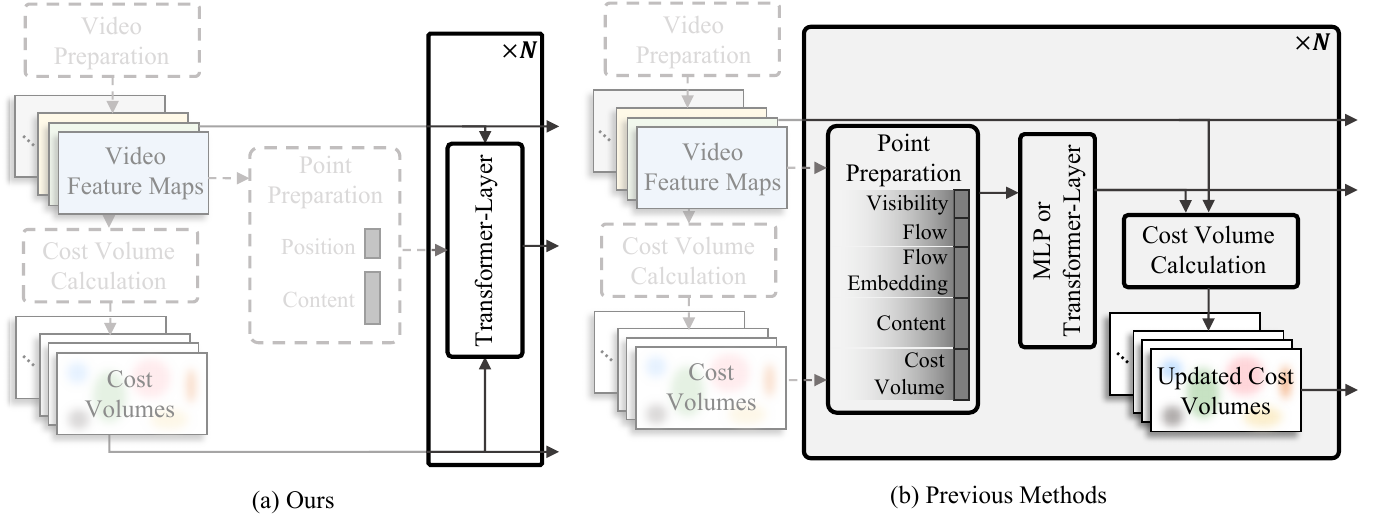}
    \caption{
    Comparison of our well-designed DETR-like simple framework with meaning-clear point modeling and previous framework with redundant designs and blackbox point modeling. The operation within the dashed box will execute only once.
    }
    \label{fig.comparison}
\end{figure*}

Compared with optical flow estimation, the most important problem that TAP needs to address is to model long-range point motion which may include occlusions along the temporal axis. 
MFT~\cite{neoral2024mft} addresses this problem by extending an off-the-shelf optical flow estimation method RAFT~\cite{teed2020raft} with the ability of visibility estimating and chaining the optical flow results to obtain the trajectory of a tracking point in the whole video. 
Although MFT has demonstrated impressive results, such an extension still lacks the capability of long-range temporal information modeling and falls short in more challenging point tracking tasks. 
To tackle this issue, several works~\cite{harley2022particle, zheng2023pointodyssey, doersch2022tap, doersch2023tapir} utilize a sliding window-based approach and let the points in different frames in the same window exchange information along the temporal axis.

However, such works usually treat each tracking point independently and ignore the correlation between points, which is an inappropriate assumption when points, for example, belong to the same object, can provide contextual information for each other based on some physical laws~\cite{yuan2023physdiff,shi2023controllable,pan2023synthesizing, chen2023humanmac,lu2023humantomato}.
To account for the correlation between tracking points, CoTracker~\cite{karaev2023cotracker} proposes to track all points simultaneously using a Transformer-based architecture. 

A crucial problem in such works is how to model tracking points. As shown in Fig.~\ref{fig.comparison} (b), in previous methods with iterative refinement~\cite{harley2022particle, zheng2023pointodyssey, karaev2023cotracker}, a tracking point is modeled as a concatenation of several features, including point flow vector as local movement, point flow embedding, point visibility, point content feature, and local correlation as cost volume. These features are normally well-designed in optical flow estimation algorithms and have clear physical meanings. However, previous methods~\cite{pips, karaev2023cotracker, zheng2023pointodyssey} simply concatenate all features and send them as a blackbox vector to MLPs~\cite{harley2022particle, zheng2023pointodyssey} or Transformers~\cite{karaev2023cotracker} and expect MLPs or Transformers to decipher and utilize the features. 
We suggest that the previous methods may not fully achieve a clean model or facilitate ease of understanding. Therefore, a conceptually simple and effective approach might be beneficial for the task.

Inspired by DEtection TRansformer (DETR)~\cite{carion2020end} and its follow-ups~\cite{ liu2022dab,li2022dn,zhang2022dino, zhu2020deformable, ren2023detrex, ren2023strong, li2023lite, liu2023detection, liu2023grounding, ren2024grounded},
we find that point tracking bears a great resemblance to object detection and tracking. In particular, in each video frame, tracking points can be essentially regarded as queries, which have been extensively studied in DETR-like algorithms~\cite{zhu2020deformable, wang2022anchor, meng2021conditional, liu2022dab, li2022dn, zhang2022dino, li2023visual}. 
With this motivation, we follow DETR-like algorithms to design a simple baseline model for tracking any point with Transformers. 

Our pipeline is illustrated in Fig.~\ref{fig.comparison} (a). 
In our framework, in every frame, each tracking point is represented as a point query, which has a clear meaning.
Each query consists of two parts, a positional part (point coordinate) and a content part, and will be refined layer by layer.
Its visibility is predicted by its updated content feature.
For multiple frames in a sliding window, queries belonging to the same tracking point can exchange information through self-attention operation along the temporal dimension. 
All such operations are common and well-designed in object detection and make the model conceptually simple yet performance-wise strong. 

We introduce some designs based on the DETR-like model to further boost performance. 
To address the difference between point tracking and object detection/tracking, we take into account the well-established cost volume~\cite{teed2020raft} into the Transformer decoder. This is driven by the fact that, compared with object detection/tracking, point tracking requires more local and low-level features to precisely locate and track desired points. 
Furthermore, we propose a simple yet effective design for updating content features within the decoder and between the windows to convey longer temporal information while mitigating the drifting issues in the context of TAP.

We conduct experiments on several challenging TAP datasets and demonstrate superior performance over prior works. 
Our model surpasses the current state of the art (CoTracker) on the DAVIS dataset under the same setting (63.0 vs. 60.7), and achieves this with 1.3 times faster speed.
Remarkably, {\methodname} outperforms CoTracker even when CoTracker deliberately tracks each single point at a time (63.0 vs. 62.2), while maintaining a 25 times faster speed.

\section{Related Work}

\noindent\textbf{Optical Flow}. Optical flow estimation is a long-standing fundamental computer vision task. Extensive research has been conducted in the past few decades~\cite{Horn_Schunck_1981, Black_Anandan_2002, Bruhn_Weickert_Schnörr_2005}. In the last ten years, deep learning-based methods~\cite{Dosovitskiy_Fischer_Ilg_Hausser_Hazirbas_Golkov_Smagt_Cremers_Brox_2015, Ilg_Mayer_Saikia_Keuper_Dosovitskiy_Brox_2017, Xu_Ranftl_Koltun_2017, Sun_Yang_Liu_Kautz_2018, Wang_Zhong_Dai_Zhang_Ji_Li_2020, Jiang_Lu_Li_Hartley_2021, Xu_Yang_Cai_Zhang_Tong_2021, Zhang_Woodford_Prisacariu_Torr_2021, Huang_Shi_Zhang_Wang_Cheung_Qin_Dai_Li, Shi_Huang_Li_Zhang_Cheung_See_Qin_Dai_Li_2023, Zhao_Zhao_Zhang_Zhou_Metaxas} have dominated this task. In particular, thanks to the regularity of image grid features, DCFlow~\cite{Xu_Ranftl_Koltun_2017} firstly utilizes cost volume for optical flow estimation. Inspired by DCFlow, many follow-up works~\cite{Sun_Yang_Liu_Kautz_2018, Wang_Zhong_Dai_Zhang_Ji_Li_2020, teed2020raft} are built upon the cost volume, validating the effectiveness and robustness of cost volume in optical flow estimation. Among the follow-ups, the paradigm of recurrently looking up the cost volume and iteratively updating the optical flow estimation proposed by RAFT~\cite{teed2020raft} obtained a remarkable performance and inspired many subsequent works~\cite{Xu_Yang_Cai_Zhang_Tong_2021, Zhang_Woodford_Prisacariu_Torr_2021}. However, optical flow only addresses the correspondence problem between two consecutive frames, which is incapable of handling the occlusion issue when a tracking point is occluded in a long-time video sequence.

\noindent\textbf{Tracking Any Point}.
Compared to optical flow, each point to be tracked in the TAP task is arbitrarily selected in a video and is required to be tracked across the entire video. This task is more general and challenging. TAP-Vid~\cite{doersch2022tap} first formalizes the task and develops a challenging benchmark for the research community. 
MFT~\cite{neoral2024mft} proposes to track points by selecting the most reliable optical flow chain. However, the inherent limitations of optical flow estimation make MFT hard to handle more challenging point tracking tasks.
OmniMotion~\cite{wang2023tracking} proposes to tackle this issue from the perspective of 3D. However, it requires a costly time-consuming test time optimization which prevents it from being applied in online scenarios. 
Recently, some general end-to-end point trackers have been proposed~\cite{pips, zheng2023pointodyssey, doersch2022tap, doersch2023tapir}, such as PIPs and TAP-Net. However, limited by the irregular data structure, they track all points in parallel independently. CoTracker~\cite{karaev2023cotracker} proposes to utilize the flexible Transformer architecture to construct the interaction between points and obtain remarkable performance. 

\section{{\methodname} Model}
\begin{figure*}[t]
    \centering
        \includegraphics[width=1\linewidth]{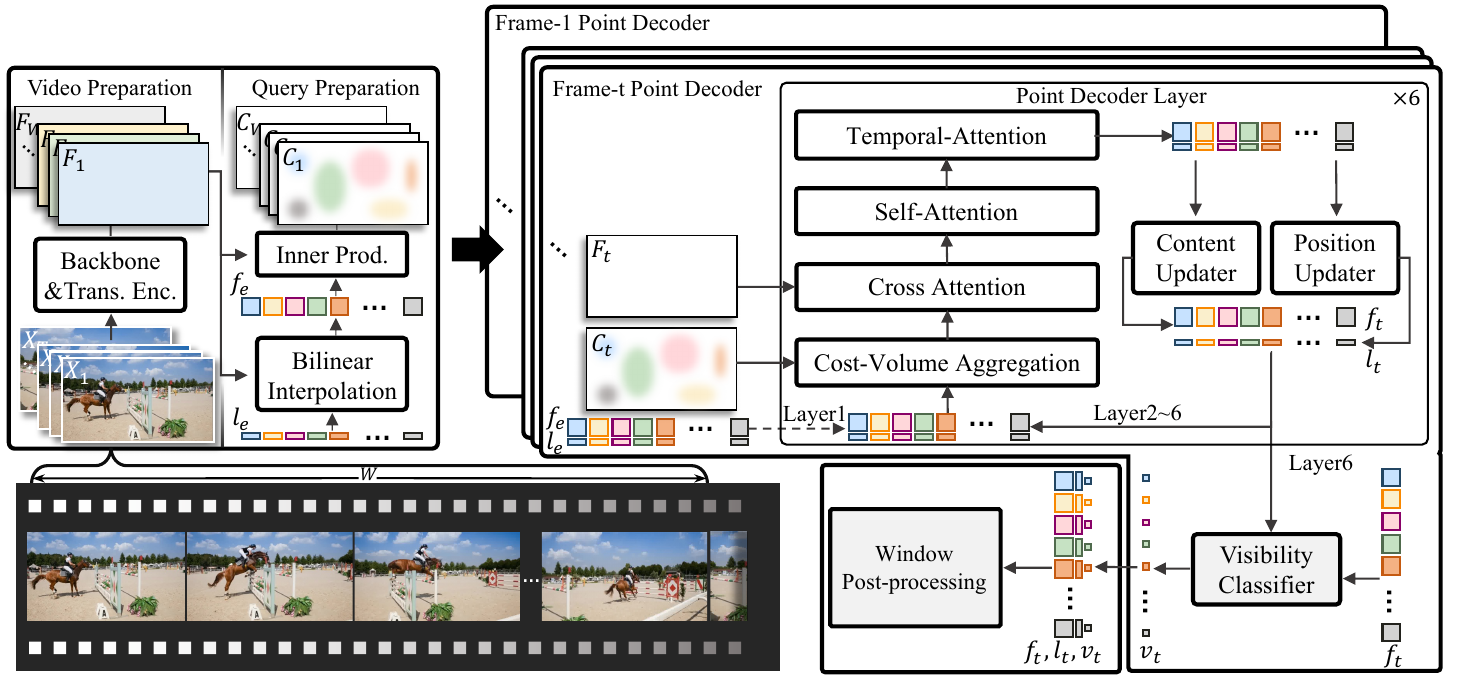}
    \caption{
    The overview of \methodname. The video preparation and query preparation parts provide the multi-scale feature map, point queries, and the cost volumes for the point decoder. The point decoder takes these elements as input and processes all frames in parallel. The outputs of the point decoder are sent to our window post-processing module to update the states of the point queries to their belonging tracking points.
    }
    \label{fig.pipeline}
\end{figure*}

\subsection{Task Definition and Overview}
\noindent\textbf{Task Definition}.
Given a video, with $T$ frames and any of the $i$-th tracking point in the video with an initial location $l_e^i=(x^i_e, y^i_e)$, our goal is to track the point across the video to obtain its trajectory, including its location sequence $L^i=\{l_t^i\}_{t=1}^T$, $l_t^i = (x_t^i,y_t^i)$, and its visibility sequence $V^i = \left\{v_t^i\right\}_{t=1}^T, v_t^i \in \{0, 1\}$. Here $e$ is a special index indicating the time stamp when the tracking point first emerges or starts to be tracked. \\
\noindent\textbf{Overview}. Our model mainly consists of four parts as shown in Fig.~\ref{fig.pipeline}. They are video preparation, query preparation, point decoder, and window post-processing. 
Following previous works~\cite{karaev2023cotracker, pips}, we use a sliding-window strategy and process $W$ frames once at a time. So for each window, 
video preparation is to extract feature maps for each frame with a backbone and a Transformer Encoder. 
Query preparation is to prepare initial locations $l_e$, content features $f_e$, and cost volumes $\left\{C_t\right\}_{t=1}^W$ for point queries in all frames of the window, 
where each point query in a frame has its unique belonging tracking point.
The point decoder takes the point queries as input to detect their belonging tracking points' states as in DETR-like methods in all frames in parallel. 
Finally, there is a window post-processing part to update the ultimate states of each point query to the trajectory of its belonging tracking point.

\subsection{Video Preparation}

Following previous Transformer-based detectors and segmentors~\cite{liu2022dab,li2022dn,zhang2022dino,zhu2020deformable} we use a convolutional neural network as our backbone to get the multi-scale image feature maps and send the image feature maps into a Transformer-encoder to further improve the quality and receptive field of the image features. 
We define the final multi-scale image feature maps for the $t$-th frame as $F_t = \{F_{t,s}\}_{s=1}^S$,
where $S$ is the number of feature scales. The feature maps of all frames are obtained independently in parallel.
\subsection{Query Preparation}
In every frame, each tracking point will be assigned with a point query. The point query belonging to the $i$-th tracking point in the $t$-th frame is responsible for detecting the most matching point of its belonging point in the $t$-th frame.

\noindent\textbf{Content Feature and Location}. 
To obtain an initial content feature for a point query that accurately describes the point to be detected, we perform bilinear interpolation on the feature maps at the location where the belonging tracking point of the query point initially appears or starts to be tracked. Thus for the point queries that belong to the $i$-th point,
their initial content feature can be obtained by
\begin{equation}
    \begin{gathered}
        f^i_{e} =\texttt{MLP}\left(\texttt{Cat}\left(\texttt{Bili}\left(F_{e^i, 1}, l^i_{e}\right), \texttt{Bili}\left(F_{e^i, 2}, l^i_{e}\right), ..., \texttt{Bili}\left(F_{e^i, S}, l^i_{e}\right)\right)\right), \\
        \forall 1\leq  t \leq T, \quad f_t^i \Leftarrow f_e^i
    \end{gathered}
\end{equation}
where $e^i$ is the timestamp when the $i$-th point first emerges or starts to be tracked. $\texttt{Bili}$ and $\texttt{Cat}$ stand for the bilinear interpolation and concatenation, respectively. $\texttt{MLP}$ is a multi-layer perceptron, which works for the fusion of point features sampled on multi-scale feature maps. 
At the same time, the initial locations of the point queries that belong to the $i$-th point are initialized as $l^i_{e}$
\begin{equation}
    \forall 1\leq  t \leq T, \quad l_t^i \Leftarrow l_e^i
\end{equation}
For notation simplicity, in the following, we denote the query that belongs to the $i$-point in the $t$-th frame as a tuple $q_t^i = [f_t^i, l_t^i]$.

\noindent\textbf{Cost Volume}. The cost volume gives us an initial visual similarity between a point query and each pixel of the image. 
The effectiveness of cost volume for feature matching has been validated in many stereo-matching~\cite{li2023stereoscene, liang2019stereo, zhao2023high, shen2021cfnet} and optical flow estimation methods~\cite{Xu_Ranftl_Koltun_2017, Jiang_Lu_Li_Hartley_2021}. Considering the characteristics of the TAP task, we further incorporate cost volume into our framework. However, different from previous works~\cite{karaev2023cotracker} that frequently recalculate the cost volume once the content feature of a query is updated, we only calculate cost volume once before the beginning of our decoder. 
This strategy keeps our decoder clean and the target of multi-layer refinement stable. 
In detail, to obtain the cost volume of the point query $q_t^i$
we conduct inner product between the content feature of the point query and the image feature maps as in previous works~\cite{Xu_Ranftl_Koltun_2017, karaev2023cotracker}
\begin{equation}
    C_{t,s}^i= \texttt{InnerProd}(F_{t,s}, f_t^i), 
\end{equation}
where $\texttt{InnerProd}$ indicates the inner product operation.  
Note that computing cost volume only once at the beginning does not mean that we do not update the cost volume anymore.
For more details about the updating of cost volume and the effect of the updating on performance, please refer to Sec.~\ref{Sec.slidingwindow} and Sec.~\ref{Sec.ablation_iter_cost_volume}. 

\subsection{Point Decoder}
Given that when focusing on a single frame, the function of a point tracker essentially involves detecting the most matched point within the image. Therefore, the components of the original decoder align naturally with the TAP task, suggesting preserving these modules, particularly the self-attention and cross-attention.

Taking into consideration the characteristics of the TAP task, we further incorporate the well-validated cost volume into our decoder through a cost volume aggregation module. 
But as we have discussed, instead of updating cost volume frequently as in prior studies~\cite{karaev2023cotracker}, to maintain the simplicity of the decoder, we treat the prepared cost volume as a static feature map similar to the original image feature maps. We directly reuse the cost volume across different layers of our decoder. 
To further utilize the temporal information, we adapt the attention mechanism along the temporal dimension through a temporal attention module.

Since each frame behaves the same in our point decoder, without loss of generality, in this section we take the $t$-th frame as an example to explain our decoder for simplicity. 

\noindent\textbf{Cost Volume Aggregation}.
Cost volume provides a basic visual similarity between the tracking point and a video frame. Instead of regressing the optical flow in a one-shot manner as in TAP-Net~\cite{doersch2022tap}, thanks to the multi-layer design in our point decoder, we follow RAFT~\cite{teed2020raft} to aggregate cost volume locally. 
More specifically, for the point query $q^i_t$, we conduct bilinear interpolation on the cost volume around its location in a grid form, this operation is commonly referred to as grid sampling $\texttt{GridSample}$. Then the sampled cost vector $c^i_t\in \mathbb{R}^{G\cdot G}$ is fused into the content feature of the point query to provide a basic perception of the image. The process can be formulated as 
\begin{equation}
    \begin{split}  
    c^i_{t,s} &= \texttt{GridSample}\left(C^i_{t,s}, \texttt{Grid}\left(l^i_t, G\right)\right), \\
    f^i_t &\Leftarrow \texttt{MLP}\left(\texttt{Cat}\left(c^i_{t,1}, c^i_{t,2}, \dots, c^i_{t,S}, f^i_t\right)\right)
    \end{split}
\end{equation}
where $C^i_{t,s}$ is the cost volume of $q^i_t$ at the $s$-th scale, $\texttt{Grid}$ is a function to generate a grid of sampling locations with grid size as $G$. Note that, like in RAFT, the sampling grid is shared in multi-scale cost volumes, resulting in a larger receptive field on smaller-scale feature maps.

\noindent\textbf{Visual Feature Enhancer}. 
As shown in previous feature-matching~\cite{xu2023iterative, liu2024global} and optical flow estimation methods~\cite{Xu_Ranftl_Koltun_2017, Jiang_Lu_Li_Hartley_2021}, supplementing cost-volume with the original image features will provide more detailed geometrical information and thus result in a more robust feature to describe the points. In DETR-like architecture, the cross-attention naturally plays this role. Following~\cite{zhu2020deformable, li2023dfa3d, zhang2022dino}, we use 2D deformable attention~\cite{zhu2020deformable} to sample the multi-scale local image feature around the tracking point and fuse the sampled image feature into the point content feature. Thus for the point query $q^i_t$, this process can be formulated as
\begin{equation}
\begin{gathered}
    g^i_{t,s} = \texttt{DFA2D}(F_{t,s}, l^i_t), \\
    f^i_t \Leftarrow \texttt{MLP}(\texttt{Cat}(g^i_{t,1}, g^i_{t,2}, \dots, g^i_{t,S}, f^i_t)),
\end{gathered}
\end{equation}
where $\texttt{DFA2D}$ indicates the 2D deformable attention operation and $g^i_{t,s}$ is the local geometrical feature for the $i$-th point in the $s$-th scale.

\noindent\textbf{Interaction Among Point Queries}.
Limited by the irregularity of data structure, most previous works choose to process every point independently. Although CoTracker~\cite{karaev2023cotracker} first proposes to use the flexible attention mechanism to complete the interaction between points, their neglect of positional embedding limits the efficiency~\cite{liu2022dab}.
Following previous DETR-like methods, we add positional embedding to encourage the points queries to pay more attention to their neighboring queries, which can provide more useful contextual information. The process can be formulated as
\begin{equation}
\label{eq.self_attention}
    p_t = \texttt{PE}\left(l_t,  \tau\right), 
    f_t \Leftarrow \texttt{Attention}\left(f_t + p_t, f_t + p_t\right).
\end{equation}
In Eq.~\ref{eq.self_attention}, $f_t\in\mathbb{R}^{N\times C}$ and $l_t\in\mathbb{R}^{N\times 2}$ indicate the content features and locations for all point queries in the $t$-th frame, where $N$ and $C$ indicate the number of points to be tracked and the number of feature channels, respectively. 
$\texttt{Attention}$ and $\texttt{PE}$ in Eq.~\ref{eq.self_attention} indicate the dense attention operation~\cite{vaswani17attention} and the sinusoidal positional encoding~\cite{carion2020end}, respectively. As the key parameter of $\texttt{PE}$, $\tau\in\mathbb{R}$ represents the temperature for positional encoding~\cite{liu2022dab}, the lower $\tau$ is, the sharper positional embedding we get. 
Considering that point detection has a higher fine-grained requirement than object detection, we lower the default value of $\tau$ down.

\noindent\textbf{Interaction Along Temporal Dimension}.
To better utilize the temporal information, we append temporal attention in our decoder. Temporal attention conducts dense attention along the temporal dimension independently for each tacking point. Thus for point queries that belong to the $i$-th point, the process can be formulated as
\begin{equation}
    f^i \Leftarrow \texttt{Attention}(f^i, f^i),
\end{equation}
where $f^i \in \mathbb{R}^{W\times C}$ indicates the content features of point queries that belong to the $i$-th point in all of the $W$ frames.

\noindent\textbf{Point Query Updating}.
Following the DETR-like methods, we update the location of each point query with the help of $\texttt{Sigmoid}$. 
As for the updating of content features, affected by the occlusion and drift, compared with the features that are updated through the above 4 blocks, the initial feature of each point query is the most reliable one. Inspired by the residual connection~\cite{he15deep}, we update the content feature of point queries in a residual mechanism. More specifically,
\begin{equation}
    \begin{gathered}
        \Delta l^i_t = \texttt{MLP}(f^i_t) , \Delta f^i_t = \texttt{MLP}(\texttt{Cat}(f^i_t, f^i_{e^i}))\\
        l^i_t \Leftarrow \texttt{Sigmoid}(\texttt{Sigmoid}^{-1}(l^i_t) + \Delta l^i_t) , f^i_t \Leftarrow f^i_t + \Delta f^i_t.
    \end{gathered}
\end{equation}
where $\texttt{Sigmoid}^{-1}$ is the inverse of sigmoid function. The updated point queries will be sent as inputs to the next layer to undergo the aforementioned blocks for $D$ times for further optimization, where $D$ indicates the number of point decoder layers. Note that, unlike the regression of location, since the classification of visibility can not be updated layer by layer and the wrong supervision of visibility may affect the prediction of location, we follow~\cite{karaev2023cotracker} to only predict the visibility at the last layer. So for the point query $q^i_t$, its visibility prediction
\begin{equation}
    \begin{split}
        v^i_t = \texttt{VisibilityClassifier}(f^i_t),
    \end{split}
\end{equation}
where $\texttt{VisibilityClassifier}$ is an $\texttt{MLP}$ ended with a $\texttt{Sigmoid}$ activate function.

Upon the completion of the final decoder layer, the updated content feature, the predicted location, and the predicted visibility of each point query will be sent to subsequent modules to update the trajectory preparing for the next window.

\begin{figure*}[t]
    \centering
        \includegraphics[width=1\linewidth]{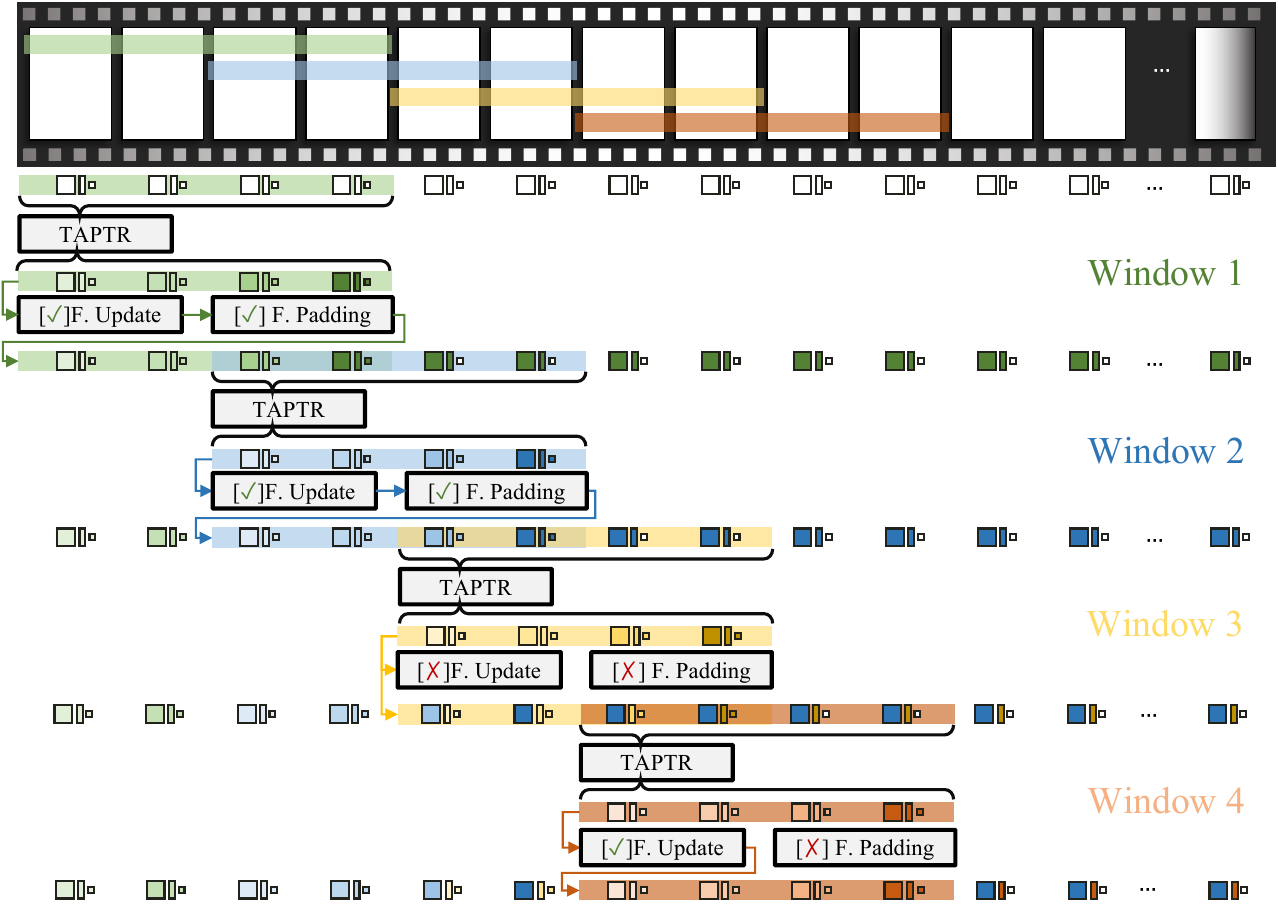}
    \caption{The overview of sliding window and window updating and padding. ``F. Update'' indicates the updating of the content feature, and ``F. Padding'' indicates the padding of the updated feature to the subsequent frames. We use window size 4 and sliding stride 2 for illustration.}
    \label{fig.slidingwindow}
\end{figure*}

\subsection{Window Post-processing}
\label{Sec.slidingwindow}
Limited by the memory limitation, we can not process all the frames of an arbitrarily long video together in parallel. Following previous works~\cite{harley2022particle, zheng2023pointodyssey, karaev2023cotracker}, we utilize the sliding window strategy to mitigate this issue. However, as illustrated in Fig.~\ref{fig.slidingwindow}, after obtaining the results of each window, our updating and padding extend beyond explicit states, including locations and visibilities, to include content features as well.

\noindent\textbf{Updating and Padding Trajectory}.
After obtaining the query results of the point queries in each frame of the current window, similar to previous works~\cite{pips, zheng2023pointodyssey,karaev2023cotracker}, we update their explicit states into their belonging trajectory.
Then based on the assumption that smaller temporal differences correspond to smaller positional differences, we pad the locations of the point queries in the last frame of the window to all corresponding point queries in the subsequent frames to reinitialize their location part. 

\noindent\textbf{Updating and Padding Content Features}.
Although the updating and padding of location transfer temporal information to subsequent windows, the lack of transferring more informative content features
loses temporal information. However, transferring the content feature without any limitation will result in a drifting problem during inference because of the inconsistency of the video length during training and inference, which will be further discussed in Sec~\ref{Sec.ablation_windowupdate}. 
To mitigate the feature drifting issue, during training and inference, instead of updating the content feature every time, we employ a random drop strategy to mitigate feature drifting issues. 
More specifically, during training, as demonstrated in Window 3 of Fig.~\ref{fig.slidingwindow}, we randomly disable feature updating, and feature padding is also disabled accordingly. The drop off of feature updating forces the network to handle the cases whether the content feature has been updated or not adaptively. 
During inference, as demonstrated in Window 4 of Fig.~\ref{fig.slidingwindow}, we keep feature updating enabled so that the information from the last window can be at least transferred to the next one. But we drop off the feature padding in a dynamic frequency according to the length of the whole video, to mitigate the accumulation of drifting in the point queries' content feature.

\noindent\textbf{Updating Cost Volume}. 
To ensure the consistency of the optimization target during the multi-layer refinement of our point decoder, we do not update the cost volume in the decoder. This strategy also keeps our decoder clean and concise. 
Since there is an overlap between two adjacent windows, frames within the overlap will be processed twice by the network. If the content feature of a point query is updated after the current window, we will update its corresponding cost volumes at the beginning of the next window. Compared to previous methods that update the cost volume at each layer, the content features at this stage are more stable, resulting in higher-quality updated cost volumes.

\subsection{Full Sequence Multi-Layer Loss}
Instead of computing loss in every window, after obtaining the whole location sequence and visibility sequence of one point, we calculate the $\texttt{L1}$ loss for the location sequence and the cross entropy loss for the visibility sequence without bells and whistles. Note that, similar to the auxiliary loss in the original DETR, we also maintain a location sequence for each layer of the point decoder. Since we only predict visibility at the last layer of the point decoder, the loss function can be formulated as
\begin{equation}
    \begin{split}
        \texttt{Loss} = \left(\omega_V \texttt{CE}(V, \tilde{V}) + \sum_{d=1}^D \omega_L \texttt{L1}(L_d, \tilde{L})\right) / N ,
    \end{split}
\end{equation}
where $\texttt{CE}$ shorts for the cross entropy loss, $D$ and $d$ are the number of decoder layers and the index of the point decoder layer respectively, $L_d$ indicates the predicted location sequences of all tracking points from the $d$-th layer, $V$ represents the predicted visibility sequence of all points from the final layer, $N$ indicates the number of points. $\tilde{L}$ and $\tilde{V}$ are the ground truth location and visibility sequences of all tracking points, respectively, $\omega_L$ and $\omega_V$ are the weights for the supervision of location and visibility.
\section{Experiments}
We conduct extensive experiments on the challenging TAP-Vid benchmark~\cite{doersch2022tap} to verify the performance of {\methodname}. Abundant ablation studies are also provided to analyze the effectiveness of each component in our framework, reflecting the adaptability of the DETR-like framework to the TAP task.
\subsection{Datasets}
For training, we follow previous methods~\cite{karaev2023cotracker} to train our model on the TAP-Vid-Kubric dataset. TAP-Vid-Kubric is a synthetic dataset consisting of 11,000 videos with 24 frames. Each video in TAP-Vid-Kubric is generated by Kubric Engine\cite{greff2022kubric} simulating the process of a set of rigid objects falling from the air and then dispersing upon impact. Each video contains annotations for tracking 2048 points. During training, we resize the resolution of the video to $512\times512$ and randomly sample 700-800 points. 
We evaluate our method on the TAP-Vid benchmark, which contains three subsets.
The first is TAP-Vid-DAVIS, which contains 30 challenging videos captured from various real scenarios with complex motion and large changes in object scale. TAP-Vid-DAVIS has about 2000 frames and 650 annotations for point tracking in total. 
The second is TAP-Vid-RGB-Stacking. Although it is a synthetic dataset, the objects in this dataset are usually texture-less, making it hard to track. 
The third one is TAP-Vid-Kinetics, which contains over 1000 labeled videos collected from YouTube.
Note that, for a fair comparison, when evaluating the benchmark, we will also downsample the video to $256\times 256$ at first. 

\subsection{Evaluation Protocol and Metrics}
\vspace{-2mm}
The TAP-Vid benchmark provides a comprehensive evaluation protocol along with metrics. 
To accommodate both the online tracker and offline trackers, TAP-Vid provides two evaluation modes. In the ``First'' mode, the tracking of a point starts from the first frame when it is visible. In the ``Strided'' mode, the tracking of a point starts from the $5$-th, $10$-th, $15$-th, $\dots$ frame, as long as it is visible in these frames. The ``Strided'' mode requires the tracker to track along bi-directions. 
TAP-Vid benchmark provides three metrics to evaluate TAP methods. Occlusion Accuracy (OA) is used to evaluate the visibility prediction. $<\delta_{avg}^x$ is the average location precision for tracking points at thresholds of 1,2,4,8,16 pixels. Average Jaccard (AJ) is a comprehensive metric reflecting the accuracy of both location and visibility.

\subsection{Implementation Details}
\vspace{-2mm}
We use ResNet50~\cite{he15deep} as our backbone and employ two layers of transformer encoder with deformable attention~\cite{zhu2020deformable}, and six layers of transformer decoder in default. 
We use the AdamW~\cite{zhuang2022understanding} optimizer and EMA~\cite{klinker2011exponential} to train our model on 8 NVIDIA A100 for about 36,000 iterations with a learning rate of 2e-4. We accumulate gradients four times using the gradient accumulation to approximate the gradients similar to a batch size of 32. We follow previous works~\cite{karaev2023cotracker} to set the window size as 8 and the stride as 4 for window sliding.

\begin{table*}[t]
\begin{center}
\resizebox{1\linewidth}{!}{ %
\begin{tabular}{l|c|ccc|ccc|ccc|ccc}
\toprule
 & PPS & \multicolumn{3}{c|}{DAVIS} & \multicolumn{3}{c|}{DAVIS-S} & \multicolumn{3}{c|}{RGB-Stacking}  & \multicolumn{3}{c}{Kinetics} \\
Method & &  AJ & $<\delta^{x}_{avg}$ & OA &  AJ & $<\delta^{x}_{avg}$ & OA &  AJ & $<\delta^{x}_{avg}$ & OA &  AJ & $<\delta^{x}_{avg}$ & OA \\
\midrule
COTR~\cite{jiang2021cotr} & -- & -- & -- & -- & 35.4 & 51.3 & 80.2& -- & -- & -- & -- & -- & -- \\
Kubric-VFS-Like~\cite{greff2022kubric} & -- & -- & -- & -- & 33.1 & 48.5 & 79.4 & -- & -- & -- & -- & -- & -- \\
RAFT~\cite{teed2020raft} & -- & -- & -- & -- & 30.0 & 46.3 & 79.6 & -- & -- & -- & -- & -- & -- \\
PIPs~\cite{harley2022particle} & -- & -- & -- & -- & 42.0 & 59.4 & 82.1 & -- & -- & -- & 31.7 & 53.7 & 72.9 \\
TAP-Net~\cite{doersch2022tap} & -- & 36.0 & 52.9 & 80.1 & 38.4 & 53.1 & 82.3 & 53.5 & 68.1 & 86.3 & 38.5 & 54.4 & 80.6 \\
MFT~\cite{neoral2024mft} & -- & 47.3 & 66.8 & 77.8 & 56.1 & 70.8 & 86.9 & -- & -- & -- & 39.6 & 60.4 & 72.7 \\
TAPIR~\cite{doersch2023tapir} & -- & 56.2 & 70.0 & 86.5 & 61.3 & 73.6 & 88.8 & \underline{55.5} & \underline{69.7} & \textbf{88.0} & \textbf{49.6} & 64.2 & 85.0 \\
OmniMotion\cite{wang2023tracking}  & -- & 52.7 & 67.5 & 85.3 & -- & -- & -- & -- & -- & -- & -- & -- & --  \\
CoTracker-Single\cite{karaev2023cotracker} & -- & 60.6 & 75.4 & \underline{89.3} & 64.8 & 79.1 & 88.7 & -- & -- & -- & 48.7 & \underline{64.3}& \textbf{86.5} \\
CoTracker2-All\cite{karaev2023cotracker} & 15.7 & 60.7 & \underline{75.7} & 88.1 & -- & -- & -- & -- & -- & -- & -- & -- & -- \\
CoTracker2-Single\cite{karaev2023cotracker} & 0.8 & \underline{62.2} & \underline{75.7} & \underline{89.3} & 65.9 & \textbf{79.4} & \underline{89.9} & -- & -- & -- & -- & -- & -- \\
\midrule
\textcolor[rgb]{0.753,0.753,0.753}{BootsTAP$^\dag$\cite{doersch2024bootstap}} & \textcolor[rgb]{0.753,0.753,0.753}{--} & \textcolor[rgb]{0.753,0.753,0.753}{61.4} & \textcolor[rgb]{0.753,0.753,0.753}{74.0} & \textcolor[rgb]{0.753,0.753,0.753}{88.4} & \textcolor[rgb]{0.753,0.753,0.753}{66.4} & \textcolor[rgb]{0.753,0.753,0.753}{78.5} & \textcolor[rgb]{0.753,0.753,0.753}{90.7} & \textcolor[rgb]{0.753,0.753,0.753}{--} & \textcolor[rgb]{0.753,0.753,0.753}{--} & \textcolor[rgb]{0.753,0.753,0.753}{--} & \textcolor[rgb]{0.753,0.753,0.753}{54.7} &  \textcolor[rgb]{0.753,0.753,0.753}{68.5} & \textcolor[rgb]{0.753,0.753,0.753}{86.3} \\
\midrule
Ours & \textbf{20.4} & \textbf{63.0} & \textbf{76.1} & \textbf{91.1} & \textbf{66.3} & \underline{79.2} & \textbf{91.0} & \textbf{60.8} & \textbf{76.2} & \underline{87.0} & \underline{49.0} & \textbf{64.4} & \underline{85.2} \\
\bottomrule
\end{tabular}
}
\caption{Comparison of {\methodname} with prior methods on TAP-Vid. Points-Per-Second (PPS) indicates how many points can be tracked across the whole video per second on DAVIS dataset on average. Note that, BootsTAP$^\dag$ is a concurrent work pre-printed in Feb.~2024 and introduces extra 15M video clips from publicly accessible videos for training.}
\vspace{-4mm}
\label{tab:benchmark_tapvid}
\end{center}
\end{table*}

\subsection{Comparison with the State of the Arts}
\label{Sec.benchmark_tapvid}
\vspace{-2mm}
We evaluate {\methodname} on the TAP-Vid~\cite{doersch2022tap} benchmark to show its superiority. As shown in Table~\ref{tab:benchmark_tapvid}, {\methodname} shows significant superiority compared with previous SoTA methods across the majority of metrics. 
Note that, The concurrent work BootsTAP~\cite{doersch2024bootstap} is trained with additional real-world data. 
To evaluate the tracking speed of different methods fairly, we compare the Point Per Second (PPS), which is the average number of points that a tracker can track across the entire video per second on the DAVIS dataset in the ``First'' mode. The results show the speed advantage of {\methodname}.

\begin{table*}[t]
\begin{center}
\resizebox{1\linewidth}{!}{ %
\begin{tabular}{c|c|c|c|c|c|c|c|c|ccc}
\toprule
Row & Small Temp.&  Trans. Enc. & Self Att. & Temp. Att. &  C. V. &  C. Attn. & Res. Up. & Win. Up. &  AJ & $<\delta^{x}_{avg}$  & OA \\ 
\midrule
1 & \checkmark & \checkmark & \checkmark & \checkmark & \checkmark & \checkmark &\checkmark &\checkmark & \textbf{63.0} & \textbf{76.1} & \textbf{91.1} \\
2 & \ding{55}  & \checkmark & \checkmark & \checkmark & \checkmark & \checkmark &\checkmark &\checkmark & 61.9 & 75.4 & 90.3 \\
3 & \ding{55}  & \ding{55}  & \checkmark & \checkmark & \checkmark & \checkmark &\checkmark &\checkmark & 60.9 & 75.2 & 88.9 \\
4 & \ding{55}  & \ding{55}  & \ding{55}  & \checkmark & \checkmark & \checkmark &\checkmark &\checkmark & 58.4 & 72.1 & 88.3 \\
5 & \ding{55}  & \ding{55}  & \ding{55}  & \ding{55}  & \checkmark & \checkmark &\checkmark &\checkmark & 51.6 & 66.7 & 84.5 \\
6 & \ding{55}  & \ding{55}  & \ding{55}  & \ding{55}  & \ding{55}   & \checkmark &\checkmark &\checkmark & 46.8 & 61.3 & 82.4 \\
7 & \ding{55}  & \ding{55}  & \ding{55}  & \ding{55}  & \checkmark  & \ding{55} &\checkmark &\checkmark & 50.0 & 65.0 & 83.4 \\
8 & \ding{55}  & \ding{55}  & \ding{55}  & \ding{55}  & \ding{55}  & \checkmark &\ding{55} &\checkmark & 45.1 & 60.0 & 82.4 \\
9 & \ding{55}  & \ding{55}  & \ding{55}  & \ding{55}  & \ding{55}  & \checkmark &\ding{55} &\ding{55} & 41.8 & 56.9 & 79.4 \\
\bottomrule
\end{tabular}
}
\caption{Ablation study of the key components in our method on DAVIS dataset. ``Small Temp.'', ``Trans. Enc'', ``Self. Att.'', ``Temp. Att.'', ``C. V.'', and ``C. Attn.'' are short for ``Small Temperature'', ``Transformer Encoder'', ``Self Attention'', ``Temporal Attention'', ``Cost Volume'', and ``Cross Attention'', respectively. ``Res. Up.'' indicates the residual updating of the point queries' content feature within the decoder, ``Win. Up.'' indicates the updating and padding of the content feature between windows.
}
\label{tab:abl_all}
\end{center}
\end{table*}

\subsection{Ablation of Key Components}
\label{Sec.ablation_keycomponent}
\vspace{-2mm}
As a baseline method, as shown in Table~\ref{tab:abl_all}, we provide extensive ablation studies to verify the effectiveness of each key component in {\methodname}, providing references for future work.

\noindent\textbf{Self Attention and Temperature in Positional Embedding}. Compared to the default large temperature in the positional encoding \cite{zhang2022dino} of self-attention within the context of object detection, reducing the temperature, especially 100 times, brings about 1.1 AJ improvement (Row 1 vs. Row 2). 
After further dropping the Self-Attention, there will also be a drop of 2.5 AJ (Row 3 vs. Row 4).
These two experiments reflect the importance of establishing connections among points and the validity of leveraging positional encoding to allocate more attention to the nearby points.

\noindent\textbf{Transformer Encoder}. After removing the Transformer Encoder, the comprehensive metric of AJ drops by about 1.0 (Row 2 vs. Row 3). A closer inspection reveals that the main performance loss stems from the accuracy of visibility estimation, indicating that increasing the quality and the receptive field of image feature maps gives a better understanding of the relationships between objects in the scene. 

\noindent\textbf{Temporal Information}. After removing the temporal attention from our decoder and the feature updating between every two windows, a significant drop of 6.8 AJ occurs (Row 4 vs. Row 5). If we further drop the updating of content feature between windows, there will be an additional drop of about 3.3 AJ (Row 8 vs. Row 9). The drops indicate the importance of temporal information. 
At the same time, removing temporal attention within the sliding window leads to a larger performance drop, indicating that short-term temporal information should be more important.

\noindent\textbf{Cost Volume Aggregation}. 
Since the cost volume provides a basic visual similarity between the tracking point and the original image, removing the cost volume from our decoder results in a significant drop of 4.8 AJ (Row 5 vs. Row 6), which aligns with the importance of cost volume in previous works.

\noindent\textbf{Cross Attention}. The dropping of cross-attention will result in a drop of 1.6 AJ (Row 5 vs. Row 7), indicating the importance of supplementing detailed visual information to cost volume from cross-attention. Note that, since cross-attention is considered as a basic perception of images in the transformer decoder, we keep cross-attention in the following ablations.

\noindent\textbf{Residual Updating}. The initial content feature of a tracking point provides a strong prior, which conveys the specific information of the point to be detected in every frame. Since the updating within the decoder is not stable enough, updating the content feature between every two decoder layers as in the original DETR may bring in noise. Replacing our residual updating with the original one results in a drop of 1.7 AJ (Row 6 vs. Row 8).

\begin{table*}[t]
\begin{center}
\resizebox{1\linewidth}{!}{ %
\begin{tabular}{c|c|c|c|c|ccc|ccc}
\toprule

Row & Update-Train & Pad-Train & Update-Inference & Pad-Inference & \multicolumn{3}{c|}{DAVIS} & \multicolumn{3}{c}{RGB-Stacking} \\
& & & & &  AJ & $<\delta^{x}_{avg}$  & OA &  AJ & $<\delta^{x}_{avg}$  & OA \\
\midrule
1 & \checkmark   & \checkmark &\checkmark   & \checkmark      & 54.7  & 73.0  & 77.3 & 23.1 & 34.3  & 38.6 \\
2 & Random Drop  & \checkmark &\ding{55}  & \ding{55} & 55.9  & 71.7  & 85.3 & 43.0 & 60.1 & 83.5 \\
3 & Random Drop  & \checkmark &Gap  & \checkmark        & 62.5  & 75.5  & 90.9 & 58.0 & 74.0 & 85.9 \\
4 & Random Drop  & \checkmark &\checkmark  & Gap           & \textbf{63.0}  & \textbf{76.1}  & \textbf{91.1} & \textbf{60.8} & \textbf{76.2} & \textbf{87.0} \\

\bottomrule
\end{tabular}
}
\caption{Ablation study of the content feature updating between windows. ``Gap'' here indicates dynamically gap the feature updating or padding.}
\label{tab:abl_windowupdate}
\end{center}
\end{table*}

\subsection{Ablation of Feature Updating Strategy Between Windows}
\label{Sec.ablation_windowupdate}
As shown in Table~\ref{tab:abl_windowupdate}, if we directly update our content feature without limitation, there will be a critical drifting problem (Row 1 vs. Row 4), especially in the RGB-Stacking dataset, where the videos are longer on average. 
As described in Sec.~\ref{Sec.slidingwindow}, during training we randomly drop off the updating of content features for the tracking points with a probability of 0.6. 
During inference, due to the length of every training video being 24, instead of random updating, we update the content feature every $T/24$ windows and gap the updating of the intermediate windows to ensure stability.
As shown in Row 3 of Table~\ref{tab:abl_windowupdate}, although this strategy loses some temporal information, it still works. To reserve more temporal information, we keep the feature updating open but drop the feature padding with the same gaps. Compared with the direct drop off of feature updating, this strategy at least keeps the temporal information between every two windows, and thus obtains the best performance as shown in Row 4.

\subsection{Ablation of Cost Volume Updating}
\label{Sec.ablation_iter_cost_volume}
To verify the advantage of our cost volume updating approach, we conduct ablation studies on the updating frequency. 
As shown in Table~\ref{tab:abl_costvolumeupdate}, if we update the cost volume after every iteration of the decoder as in previous methods~\cite{karaev2023cotracker}, there will be a decline of about 1.1 AJ on the DAVIS dataset and 2.8 AJ on the RGB-Stacking dataset. 
On the other hand, if we keep the cost volume never updated, although achieving comparable performance on the DAVIS dataset, the performance on the RGB-Stacking dataset still decreases by about 2.5 AJ. This is reasonable because, in this case, the cost volume can not benefit from the long temporal information, leading to worse performance on longer videos.

\begin{table}[t]
  \centering
  \begin{minipage}[b]{0.30\textwidth}
    \centering
    \resizebox{1\linewidth}{!}{ %
    \begin{tabular}{c|ccc}
    \toprule
    Supervision &  AJ & $<\delta^{x}_{avg}$  & OA \\
    \midrule
    Last Layer Only    & 55.4  & 70.0  & 87.3 \\
    All Layers     & \textbf{63.0}  & \textbf{76.1}  & \textbf{91.1} \\
    
    \bottomrule
    \end{tabular}
    }
    \caption{Ablation study of the multi-layer losses.}
    \label{tab:abl_loss}
  \end{minipage}
  \hfill
  \begin{minipage}[b]{0.25\textwidth}
    \centering
    \resizebox{1\linewidth}{!}{ %
    \begin{tabular}{c|ccc}
    \toprule
    \# Decoder &  AJ & $<\delta^{x}_{avg}$  & OA \\
    \midrule
    2              & 58.2  & 72.3  & 89.0 \\
    4              & 61.6  & 75.0  & 89.8 \\
    6              & \textbf{63.0}  & \textbf{76.1}  & \textbf{91.1} \\
    
    \bottomrule
    \end{tabular}
    }
    \caption{Ablation of decoder layer.}
    \label{tab:abl_numdec}
  \end{minipage}
  \hfill
  \begin{minipage}[b]{0.43\textwidth}
  \resizebox{1\linewidth}{!}{ %
    \begin{tabular}{c|ccc|ccc}
    \toprule
    & \multicolumn{3}{c|}{DAVIS}     & \multicolumn{3}{c}{RGB-Stacking} \\
    Updating Freq. & AJ & $<\delta^{x}_{avg}$ & OA  &  AJ & $<\delta^{x}_{avg}$ & OA \\
    \midrule
    Per Iter.      & 61.9  & 75.2  & 90.7 & 58.0 & 75.6 & 84.0 \\
    No  Update     & 63.0  & 76.1  & 90.9 & 58.3 & 76.0 & 84.1 \\
    Per Wind.      & \textbf{63.0}  & \textbf{76.1}  & \textbf{91.1} & \textbf{60.8} & \textbf{76.2} & \textbf{87.0} \\
    
    \bottomrule
    \end{tabular}
    }
    \caption{Ablation study of cost volume updating frequency.}
    \label{tab:abl_costvolumeupdate}
  \end{minipage}

\end{table}

\subsection{Ablation of Decoder Layer Number and Multi-layer Supervision}
\label{Sec.ablation_num_encdec}
Consistent with the conclusions drawn from previous DETR-based object detection methods~\cite{carion2020end,liu2022dab,li2022dn,zhang2022dino} and multi-layer refinement-based TAP methods~\cite{pips, zheng2023pointodyssey, karaev2023cotracker}, as shown in Table~\ref{tab:abl_numdec}, as we increase the number of decoder layers, the number of refinement increases and the performance improves accordingly.
However, if we only increase the number of decoder layers without supervising the output of each layer, the performance is still poor and even worse than the one with only 2 decoder layers, as shown in Table~\ref{tab:abl_loss}. This indicates that, if not fully supervised, the multi-layer refinement brings a negative impact instead.

\section{Visualization}
\begin{figure*}[h]
    \centering
        \includegraphics[width=1\linewidth]{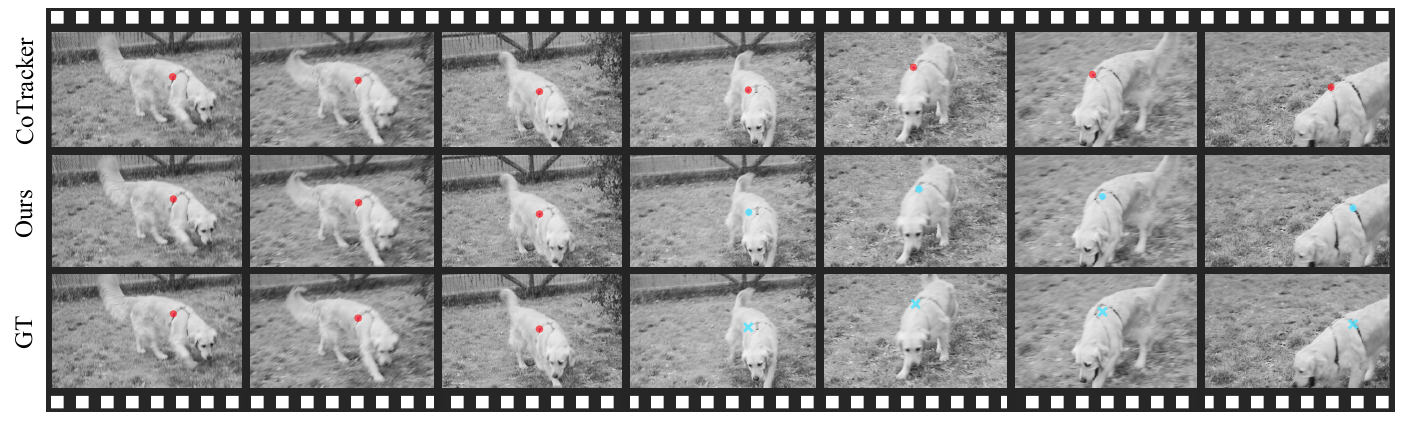}
    \caption{Red and blue indicate visible and occluded respectively. We manually supplement the ground truth location of invisible points with blue crosses for better comparison. Best view in electronic version.}
    \label{fig.vis_comp}
\end{figure*}

As shown in Fig.~\ref{fig.vis_comp}, 
when the dog turns around, CoTracker shows a significant drifting, where the tracking result shifts from the right side to the top of the dog. In contrast, {\methodname} tracks stably even when the tracking target is occluded.
For more comparisons, fancy visualizations, and corresponding videos, please refer to Sec.~\ref{sec.supp_more_compare} and Sec.~\ref{sec.supp_video_editing} in appendix.

\section{Conclusion}
In this paper, we have proposed a conceptually simple and strong baseline for tracking any point task. With the help of our designs in mitigating feature drifting we obtain SoTA performance while demonstrating an advantage in inference speed. We also conduct extensive ablations to provide references for future work. \\
\textbf{Limitations and Future work}. Due to the difficulty in annotating in the TAP task, most training data currently used is synthetic. How to further leverage detection and segmentation annotations to assist TAP tasks is an interesting problem remaind to be explored in our future work.
\onecolumn
\section*{Appendix}

\appendix

\section{BADJA Benchmark}
BADJA is a dataset of animal joint tracking, including 7 videos from the DAVIS dataset and 2 extra ones. Since it approximately fivefold downsamples the video, the movement of the objects in the video is very fast, and the movements of the target tracking points are also intense accordingly, making them hard to track. 
Although seven of them are collected from DAVIS, the videos' FPS and annotations are different from the TAP-Vid-DAVIS dataset.

To provide a more detailed comparison, follow both the settings in PIPs and CoTracker. 
The first one is the ``$\delta^{seg}$'' one following PIPs. This setting only considers the 7 videos collected from DAVIS and calculates the proportion of points with discrepancies less than $0.2\sqrt{A}$ compared to the ground truth (GT) as its metric, where $A$ indicates the area of the tracking animal's mask. Also, this setting requires comparing the performance of each video in detail.
The second one is the ``$\delta^{3px}$'' one proposed by CoTracker. This setting considers all of the 9 videos and calculates the proportion of points with discrepancies less than 3 pixels compared to the GT as its metric. This setting requires comparing the overall performance.

As shown in Table~\ref{tab:badja_test}, {\methodname} obtains the best overall performance on both settings. 
Note that, due to the difference between the requirement of tracking physical surface points in TAP, and the requirement of tracking joints inside the animal in BADJA benchmark, the performance on BADJA can only be considered as a reference.

\begin{table*}[h]
\begin{center}
\caption{Avg.-7 indicates the average performance of the 7 videos collected from DAVIS. Avg.-All indicates the average performance of all videos, including the extra 2.}\label{tab:badja_test}
\begin{tabular}{l|cccccccc|c}
\toprule
{Method} & \multicolumn{8}{c|}{$\delta^{seg}$}  & $\delta^{3px}$ \\
& bear & camel & cows & dog-a & dog & horse-h & horse-l & Avg.-7 & Avg.-All \\
\midrule
DINO~\cite{dino} & 75.0 & 59.2 & 70.6 & 10.3 & \underline{47.1} & 35.1 & 56.0 & 50.5 & --\\
ImageNet ResNet~\cite{he16deep} & 65.4 & 53.4 & 52.4 & 0.0 & 23.0 & 19.2 & 27.2 & 34.4 & -- \\
CRW~\cite{jabri2020space} & 66.1 & 67.2 & 64.7 & 6.9 & 33.9 & 25.8 & 27.2 & 41.7 & -- \\
VFS~\cite{xu2021vfs} & 64.3 & 62.7 & 71.9 & 10.3 & 35.6 & 33.8 & 33.5 & 44.6 & -- \\
MAST~\cite{lai2020mast} & 51.8 & 52.0 & 57.5 & 3.4 & 5.7 & 7.3 & 34.0 & 30.2 & -- \\
RAFT~\cite{raft} & 64.6 & 65.6 & 69.5 & 13.8 & 39.1 & 37.1 & 29.3 & 45.6 & 7.6 \\
PIPs~\cite{pips} & \underline{76.3} & \underline{81.6} & \underline{83.2} & \textbf{34.2} & 44.0 & \textbf{57.4} & \underline{59.5} & \underline{62.3} & 13.5 \\ 
TAP-Net~\cite{tapvid} & -- & -- & -- & -- & -- & -- & -- & -- & 6.3 \\
TAPIR~\cite{doersch2023tapir} & -- & -- & -- & -- & -- & -- & -- & -- & 15.2 \\
CoTracker~\cite{karaev2023cotracker} & -- & -- & -- & -- & -- & -- & -- & -- & \underline{18.0} \\
\midrule
\methodname (ours) & \textbf{81.8} & \textbf{86.8} & \textbf{89.8} & \underline{26.9} & \textbf{52.6} & \underline{47.1} & \textbf{63.2} & \textbf{64.0} & \textbf{18.2} \\
\bottomrule
\end{tabular}
\end{center}
\end{table*}
\vspace{-7mm}

\section{{\methodname} in Trajectory Prediction}
\vspace{-3mm}
\label{sec.supp_traj}
Here we show the handwriting trajectory prediction of {\methodname} for an example. For more details please refer to the videos in our  code repository. 
Corresponding video names are provided in Sec.~\ref{sec.video_correspondences}.

\begin{figure*}[bhtp]
    \centering
        \includegraphics[width=0.83\linewidth]{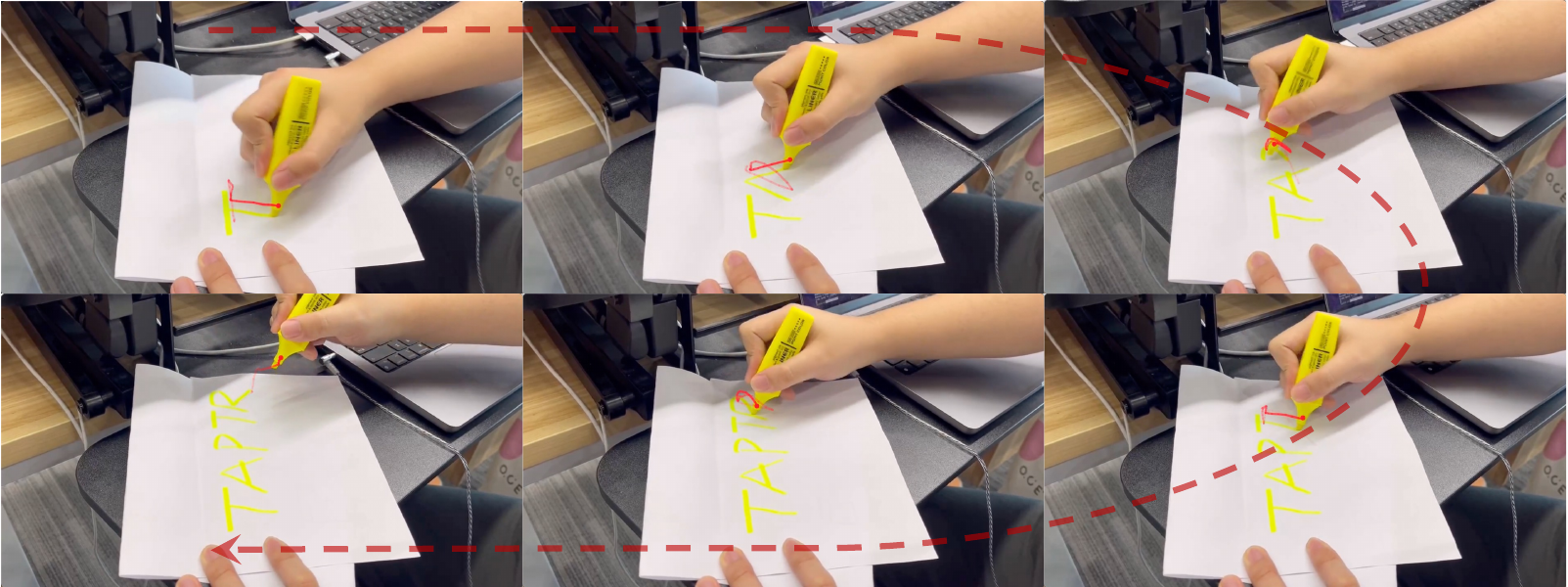}
    \vspace{-2mm}
    \caption{
    The trajectory of handwriting predicted by {\methodname}.
    }
    \label{fig.trajectory}
\end{figure*}

\newpage

\section{{\methodname} in Video Editing}
\vspace{-3mm}
\label{sec.supp_video_editing}
Here we show the results of the video editing using {\methodname}. We sample points in the editing area of the first frame and track these points across the whole video. For more details please refer to the videos in our code repository. 
Corresponding video names are provided in Sec.~\ref{sec.video_correspondences}.

\begin{figure*}[bhtp]
    \centering
        \includegraphics[width=0.75\linewidth]{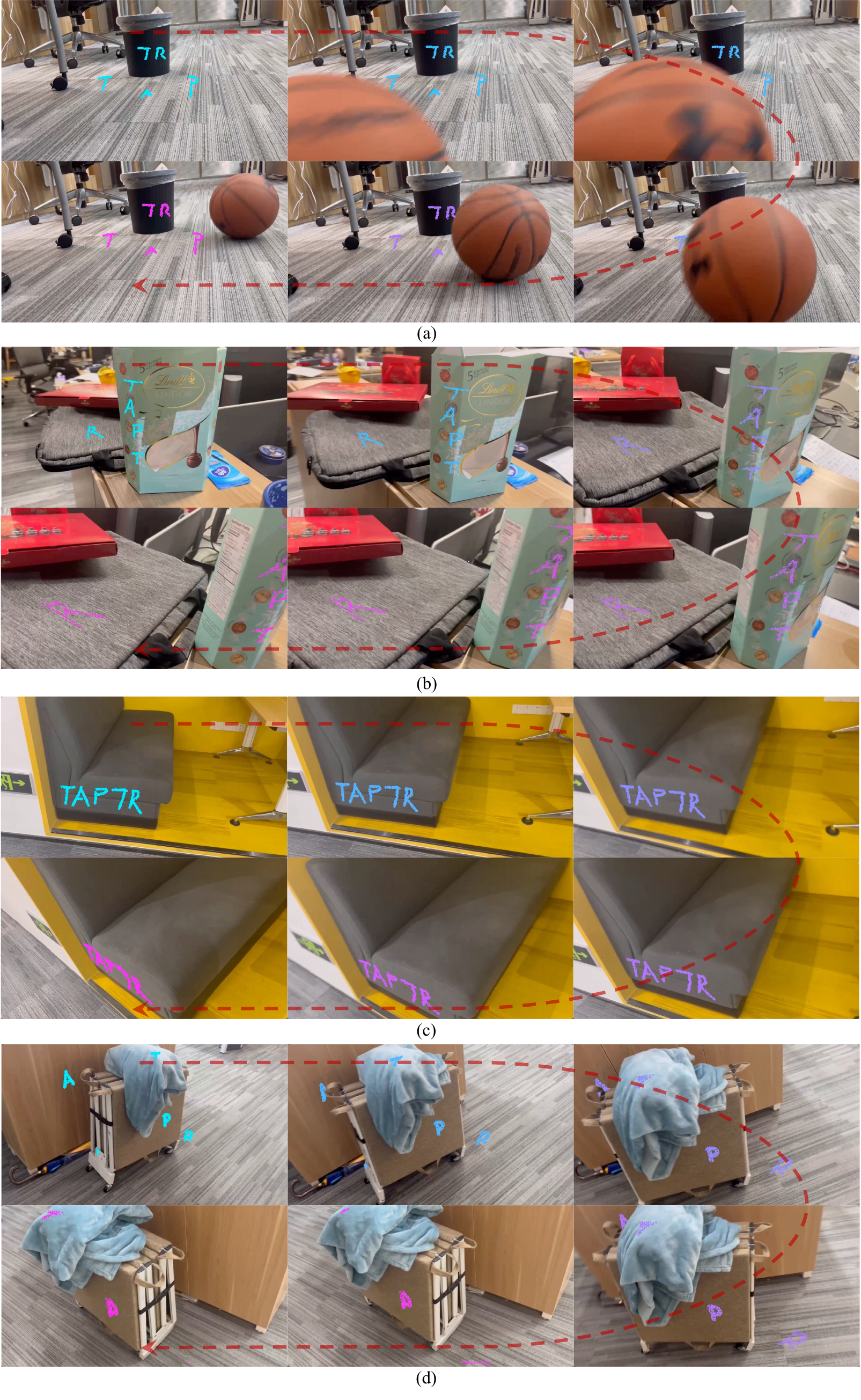}
    \vspace{-2mm}
    \caption{
    Edit video with {\methodname}. The color of the editing area changes over time.
    }
    \vspace{-30mm}
    \label{fig.fancy_visualization}
\end{figure*}

\newpage
\section{More Comparisons}
\vspace{-3mm}
\label{sec.supp_more_compare}
Here we show more comparisons between the current state-of-the-art method and {\methodname}. For more details please refer to the videos in our code repository. 
Corresponding video names are provided in Sec.~\ref{sec.video_correspondences}.

\begin{figure*}[bhtp]
    \centering
        \includegraphics[width=0.75\linewidth]{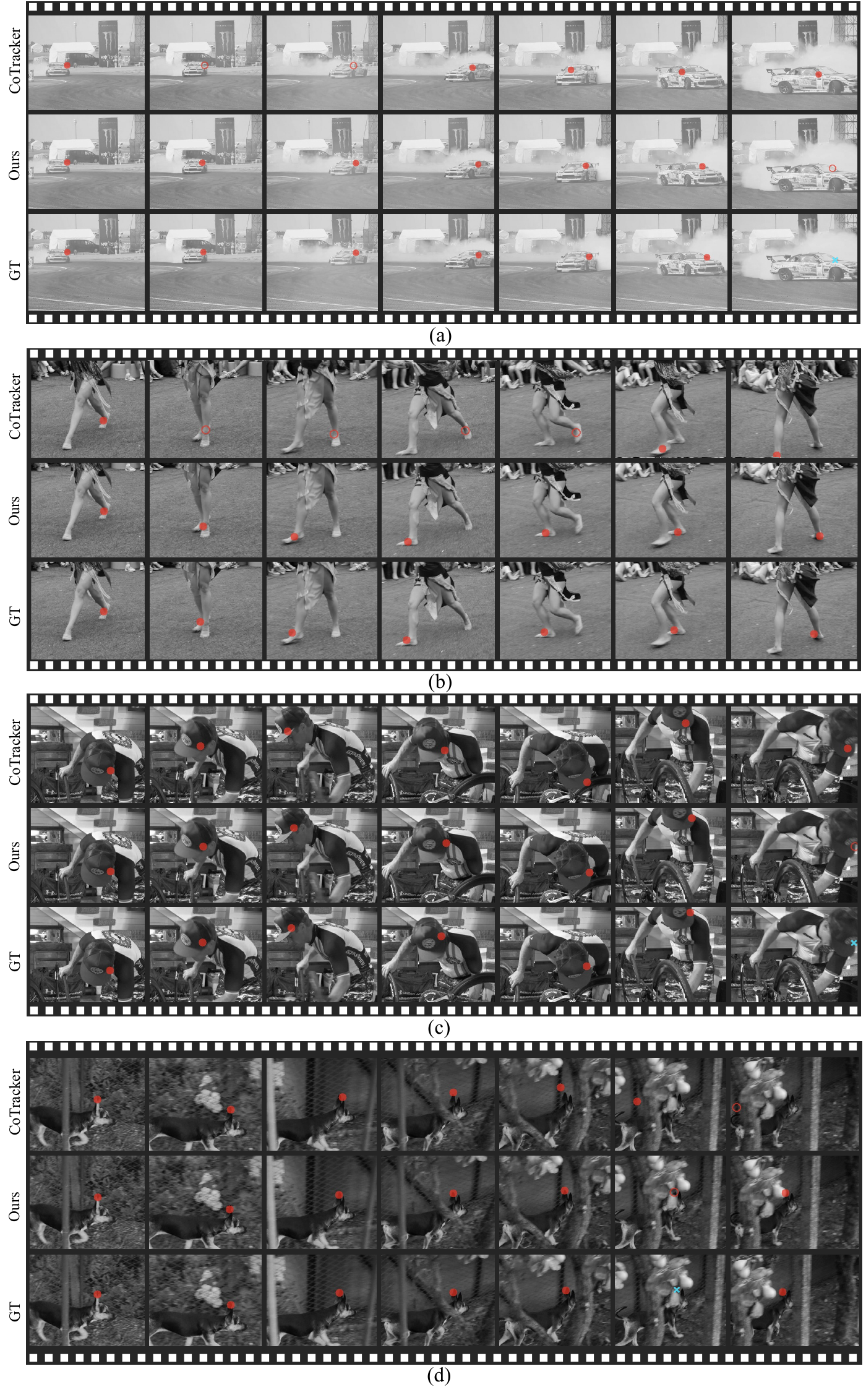}
    \vspace{-2mm}
    \caption{
    The solid and hollow red circles represent the predicted visible and invisible points. We manually supplement the GT locations of invisible points for better comparison. Best view in electronic version
    }
    \vspace{-30mm}
    \label{fig.more_comparison}
\end{figure*}

\newpage
\section{Correspondences Between Visualizations and Videos}
\label{sec.video_correspondences}
\noindent Fig.~\ref{fig.vis_comp} --> Compare\_dog\_walk.mp4 \\
\noindent Fig.~\ref{fig.trajectory} --> Traj\_Handwriting.mp4 \\
\noindent Fig.~\ref{fig.fancy_visualization} (a) --> VideoEdit\_roll\_basketball.mp4 \\
\noindent Fig.~\ref{fig.fancy_visualization} (b) --> VideoEdit\_box.mp4 \\
\noindent Fig.~\ref{fig.fancy_visualization} (c) --> VideoEdit\_sofa.mp4 \\
\noindent Fig.~\ref{fig.fancy_visualization} (d) --> VideoEdit\_bed.mp4 \\
\noindent Fig.~\ref{fig.more_comparison} (a) --> Compare\_drift.mp4 \\
\noindent Fig.~\ref{fig.more_comparison} (b) --> Compare\_dancer.mp4 \\
\noindent Fig.~\ref{fig.more_comparison} (c) --> Compare\_bike.mp4 \\
\noindent Fig.~\ref{fig.more_comparison} (d) --> Compare\_dog\_run.mp4 \\

\newpage

\bibliographystyle{splncs04}
\bibliography{egbib,main}
\end{document}